\documentclass[letterpaper, 10 pt, conference]{ieeeconf}  \IEEEoverridecommandlockouts                                                                                        

\makeatletter

\usepackage{times}  \usepackage{helvet}  \usepackage{courier}  \usepackage{url}  \usepackage{graphicx}  \usepackage{dsfont}

\usepackage{xspace}
\usepackage{times}
\usepackage{amsmath,amssymb,amsfonts}
\usepackage{verbatim}
\usepackage{graphicx}
\usepackage{psfrag,graphicx,epsfig,epsf}
\usepackage{graphics}
\usepackage{latexsym}
\usepackage{fancyhdr}
\usepackage{setspace}
\usepackage{color}
\usepackage{url}
\usepackage{mathtools}
\usepackage{amsmath,amsfonts,amssymb,mathrsfs}
\usepackage{psfrag,graphicx,epsfig,epsf}
\usepackage[ruled,linesnumbered, noend]{algorithm2e}
\usepackage{fancyhdr}
\usepackage{wrapfig}
\usepackage{subfigure}
\usepackage{mathtools}
\usepackage{url}
\usepackage{color}
\usepackage{verbatim}
\usepackage{xspace}

\title{\LARGE \bf Anytime Multi-arm Task and Motion Planning\\ for
  Pick-and-Place of Individual Objects via Handoffs}

\author{Rahul Shome and Kostas E. Bekris\thanks{The authors are with the Computer Science Department of Rutgers
  University in New Brunswick, NJ, USA. {\tt\small \{rahul.shome,
    kostas.bekris\}@rutgers.edu}}
}

\newcommand{\pose}{p}

\newcommand{\graph}{\mathcal{G}}
\newcommand{\nodes}{\mathcal{V}}
\newcommand{\node}{{v}}
\newcommand{\edges}{\mathcal{E}}

\newcommand{\prmstar}{{\tt PRM$^*$}}

\newcommand{\rrtstar}{{\tt RRT$^*$}}

\newenvironment{myitem}{\begin{list}{$\bullet$}
{\setlength{\itemsep}{-0pt}
\setlength{\topsep}{0pt}
\setlength{\labelwidth}{0pt}
\setlength{\leftmargin}{10pt}
\setlength{\parsep}{-0pt}
\setlength{\itemsep}{0pt}
\setlength{\partopsep}{0pt}}}{\end{list}}

\newtheorem{definition}{\bf Definition}

\newcommand{\dof}{{\tt DoF}}

\newcommand{\mmgraph}{\ensuremath{\mathbb{G}}}
\newcommand{\mmgimp}{\hat\mmgraph}

\newcommand{\tree}{\ensuremath{\mathbb{T} \ }}

\newcommand{\drrtstar}{\ensuremath{{\tt dRRT^*}}}

\newcommand{\cost}{\textup{cost}}

\newcommand{\cfree}{\ensuremath{\cspace^{{\rm free}}}  }
\newcommand{\tfree}{\ensuremath{\taskspace^{{\rm free}}}  }

\newcommand{\cobs}{\ensuremath{\cspace^{{\rm obs}}}  }

\newtheorem{assumption}{Assumption}

\newcommand{\object}{o}

\newcommand{\taskspace}{\mathcal{T}}

\newcommand{\arm}{m}

\newcommand{\state}{q}

\newcounter{model}

\definecolor{darkgreen}{RGB}{30,150,30}

\newcommand{\cspace}{\mathcal{C}}

\newcommand{\mode}{{{\node_{\modegraph}}}}

\newcommand{\sethree}{\mathtt{SE(3)}}
\newcommand{\posespace}{\mathcal{P}}
\newcommand{\grasp}{\mathit{g}}
\newcommand{\State}{Q}
\newcommand{\pdh}{{\tt PHP}}
\newcommand{\tamp}{{\tt TAMP}}
\newcommand{\modegraph}{\mathcal{M}}

\newcommand{\vnear}{\ensuremath{V^{\textup{near}}}}
\newcommand{\vnew}{\ensuremath{V^{\textup{new}}}}
\newcommand{\vlast}{\ensuremath{V^{\textup{last}}}}

\newcommand{\ioracle}{\mathbb{I}_\modegraph}

\newcommand{\heuristic}{\ensuremath{\mathbb{H}}}

\newcommand{\mmdrrtstar}{{\tt mm\drrtstar}}
\newcommand{\modeinit}{\node_\modegraph^{init}}
\newcommand{\modegoal}{\node_\modegraph^{goal}}
\newcommand{\modenbor}{\mathcal{N}_\modegraph}
\newcommand{\tensorset}{\mmgimp}

\newcommand{\Snear}{\ensuremath{\State^{\textup{near}}}}
\newcommand{\Snew}{\ensuremath{\State^{\textup{new}}}}

\newcommand{\mnear}{\ensuremath{\mode^{\textup{near}}}}
\newcommand{\mnew}{\ensuremath{\mode^{\textup{new}}}}

\newcommand{\snear}{\ensuremath{\state^{\textup{near}}}}
\newcommand{\snew}{\ensuremath{\state^{\textup{new}}}}

\begin{document}

\maketitle
\thispagestyle{empty}
\pagestyle{empty}

\begin{abstract}

Automation applications are pushing the deployment of many high
\dof\ manipulators in warehouse and manufacturing environments. This
has motivated many efforts on optimizing manipulation tasks involving
a single arm. Coordinating multiple arms for manipulation, however,
introduces additional computational challenges arising from the
increased \dof s, as well as the combinatorial increase in the
available operations that many manipulators can perform, including
handoffs between arms. The focus here is on the case of pick-and-place
tasks, which require a sequence of handoffs to be executed, so as to
achieve computational efficiency, asymptotic optimality and practical
anytime performance.  The paper leverages recent advances in
multi-robot motion planning for high \dof\ systems to propose a novel
multi-modal extension of the \drrtstar\ algorithm. The key insight is
that, instead of naively solving a sequence of motion planning
problems, it is computationally advantageous to directly explore the
composite space of the integrated multi-arm task and motion planning
problem, given input sets of possible pick and handoff configurations.
Asymptotic optimality guarantees are possible by sampling additional
picks and handoffs over time.  The evaluation shows that the approach
finds initial solutions fast and improves their quality over time. It
also succeeds in finding solutions to harder problem instances
relative to alternatives and can scale effectively as the number of
robots increases.

\end{abstract}

\section{Introduction}
\label{sec:motivation}
The use of manipulators for pick-and-place tasks in automation
environments is driving a variety of
applications \cite{Correll:2016aa,shome2019towards}. Coordinating more
than one manipulator brings the promise of faster execution and
enables a richer set of capabilities \cite{akbari2018combined,
shome2018rearrangement}. The caveat is that manipulators are already
high \dof\ robots, and coordinating multiple manipulators at the task
planning level \cite{Koga:1994fk, dobson2015planning} involves
searching an even larger configuration space
($\cspace$-space). Furthermore, there is a larger set of operations
beyond just picking and placing at the goal configuration, which
involve handing off an object or placing it stably so that another
manipulator grasps it.

\begin{figure}[t]
	\centering
	\includegraphics[width=0.48\textwidth]{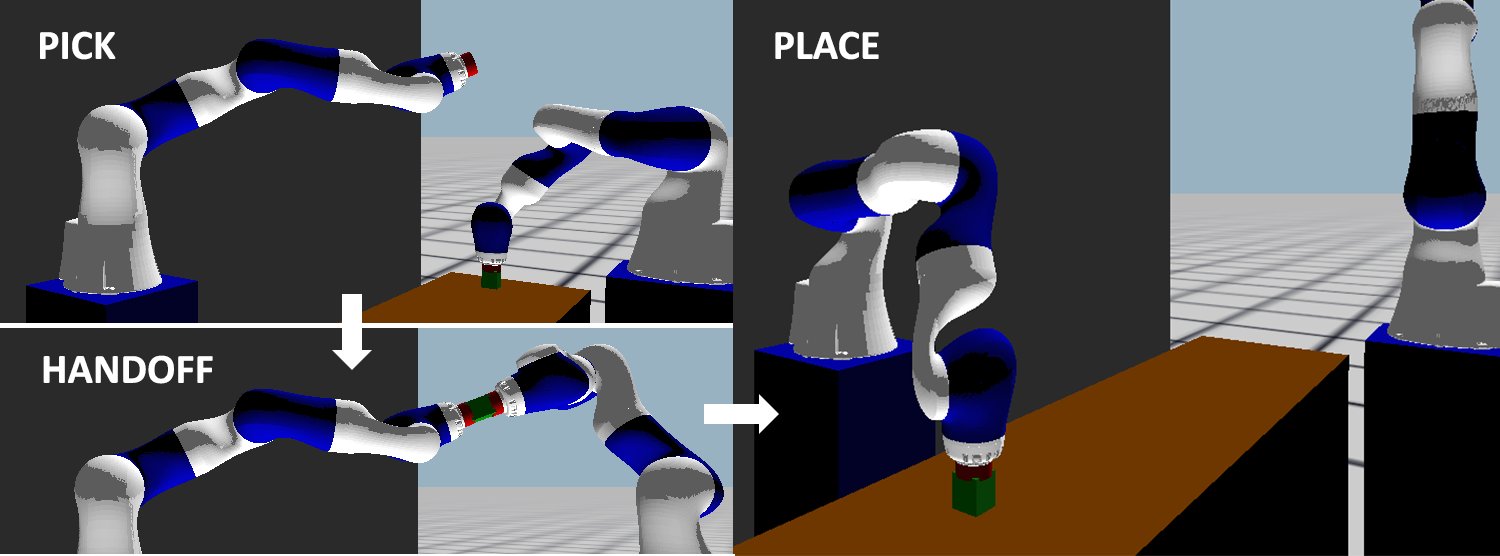}
	\vspace{-0.25in}
\caption{A pick-and-place via handoff task involving two 7 \dof\ Kuka arms and one object. The three frames show the instants of pick, handoff and then place. This extends the reachability of each robotic system.}
	\vspace{-0.25in} \label{fig:example_two_arm}
\end{figure}

Inspired by these tasks and challenges, this work aims to provide
tools for integrated task and motion planning involving multiple
manipulators with performance guarantees, such as asymptotic
optimality, and practical computational performance, such as anytime
behavior. Towards this objective, the current paper focuses on the
case of pick-and-place tasks involving individual objects, which
require handoffs between at least two manipulators, as in
Fig \ref{fig:example_two_arm}. Handoffs significantly enhance the
reachability capabilities of a static automation infrastructure beyond
the volume of the workspace reachable by a single manipulator.

\subsection{Foundations}
\vspace{-.05in}

The pick-and-place via handoffs problem, abbreviated here as \pdh,
exposes most of the prototypical challenges of both multi-robot
planning and integrated task and motion planning (\tamp). 
The search for a multi-arm \tamp\ solution requires adequate exploration of both different task operations, and motions, while allowing adequate opportunity to refine both components in terms of solution quality. This motivates anytime, integrated, multi-arm \tamp\ planning which poses unique challenges, aspects of which have been addressed in previous work.

Multi-robot motion planning research has approached the problem with decentralized solutions~\cite{ErdLoz86,GhrOkaLav05,LavHut98b,PenAke02,Berg:2005bh,Berg:2009ve}
which reduce search space size by partitioning the problem into several
sub-problems but typically lack
completeness and optimality guarantees.  
In contrast, centralized approaches
\cite{kh-pppi05,OdoLoz89,shh12,SoloveySH16:ijrr,Svestka:1998ud,Wagner:2015bd}
usually work in the combined high-dimensional configuration spaces, and
thus tend to be slower than decoupled techniques but provide stronger theoretical guarantees.

Sampling-based approaches have been at the forefront of
motion planning research ~\cite{Kavraki1996Probabilistic-R, LaValle2001}, including more
recent asymptotically optimal (AO)
variants~\cite{Karaman2011Sampling-based-,
janson2015fast}.
AO methods, however, require
building rather dense data structures and can suffer scalability
issues when dealing with multi-robot problems, where the number
of \dof s of the system increases. 
Recent advances in sampling-based multi-robot motion planning focused on high \dof\ systems,
such as the \drrtstar\ method \cite{Dobson:2017aa, shome2019drrt},
help deal with the increase in the size of the composite
$\cspace$-space by automatically taking advantage of any natural
decoupling present in the problem and precomputation~\cite{Svestka:1998ud,Wagner:2015bd,SoloveySH16:ijrr} expressing each
robot's reachability region. This work aims to build on top of this
anytime, AO, sampling-based approach.

\tamp\ is a well studied problem in
robotics. Work
has focused on describing the general
formulation \cite{simeon2004general} and the planning structures in
the space of tasks \cite{Simeon:2004tg,
Hauser2011Randomized-Multi-Modal-, Hauser:2010qf}, with different variants of solving tasks using manipulators including movable obstacles~\cite{Stilman:2004qa}, non-prehensile~\cite{Dogar:2011ve, Dogar22012} or optimization~\cite{toussaint2015logic} problems. 

Most of the work has focused on hierarchical strategies, which typically commit to solutions from time-budgeted underlying motion planning subroutines, and heuristically~\cite{akbari2018combined, vega2018admissible,
garrett2018ffrob} guide the search over actions in the
task space \cite{gravot2005asymov, dantam2016incremental,
Kaelbling:2011gb}. 
An important aspect of manipulation task planning is the
generation and evaluation of grasps that form these mode transitions. This is not a focus of the present work and it should be noted that we focus on devising effective search strategies for the \pdh\ problem over a pre-specified set of picks, handoffs and drops.

It should be noted that as the number of robots increases in \tamp, so does the number of modes in the search space of task planning~\cite{dobson2015planning,Harada2014A-Manipulation}. The multi-arm problem has also been explored using a heuristic search-based
approach\cite{cohen2015planning}, and using dynamical
reasoning\cite{sina2016coordinated}. 
A keen observation in previous work~\cite{vega2016asymptotically, garrett2017sample} about the nature of constraints that typically arise in \tamp\ lets these methods \textit{factorize} the search space. The key idea~\cite{garrett2017sample} is that constraints that describe modes and transitions for tasks often affect only a subset of the systems (robots or objects), and can often be dealt with independently to speed up the search.  For instance the picks only describe
a constraint on the picking arm and the object, and not other arms. This means that task constraints that express modes of the task can underspecify the configuration of robots that are not involved.

Quite recently, there has been some interest in looking at the \tamp\
problem in a more integrated fashion \cite{schmitt2017optimal,
vega2016asymptotically} and providing optimality guarantees. The benefit here is that an integrated approach ensures different choices of task modes and associated motions can be simultaneously explored and refined over time.

\subsection{Contribution}

The key insight of this work is that in the case of the
multi-arm \tamp\ problem, the principle of $ \cspace $-space decomposition and search in \drrtstar\ also expresses the factored nature of the multi-arm \tamp\ problem, thereby enabling the design of an anytime, efficient multi-arm \tamp\ planner for searching over a set of picks and handoffs. The key contributions are listed as follows:
\textit{(1) }The current work extends the principle of decomposition of the composite multi-arm space to the \tamp\ domain in the \pdh\ problem, and derives the same scalability, robustness and optimality benefits;
\textit{(2) }The proposed approach can operates over factored constraints can affect a subset of the arms, and ensures exploration of different realizations of the underspecified arm.
\textit{(3) }The method utilizes heuristics to guide the search towards fast initial solutions and the sampling-based exploration ensures anytime properties that refine the solution over time;
\textit{(4) }Experiments indicate that the method effectively scales to find solutions for a large number(demonstrated up to 5) of high \dof\ manipulators;
\textit{(5) }Asymptotic optimality arguments can be made when the set of picks and handoffs is augmented over time.

\section{Problem Setup}
\label{sec:problem}
To simplify description, the problem setup will be outlined for two
manipulators, $ \arm_1$ and $\arm_2 $, operating in a shared
workspace. The extension to more arms is straightforward and
considered in the experimental section of this paper.

Each manipulator has its own $ d_i $-dimensional $\cspace$-space
$ \cspace_{\arm_1}, \cspace_{\arm_2}$. The composite configuration
space of both arms is $ \cspace
= \cspace_{\arm_1} \times \cspace_{\arm_2} \subset \mathbb{R}^{d_1+d_2}
$. Then, a composite configuration is $ \State=(\state^1,\state^2) $,
where $ \state^i \in \cspace_{\arm_i} $.  $ \cobs \subset \cspace $ is
the obstacle subset, where either a manipulator collides with itself
or with static obstacles, or manipulators collide with each other.
$ \cfree = \cspace \setminus \cobs $ is the valid subset.

The workspace also contains a single rigid body, $ \object $, which
can attain poses in the space $ \posespace_\object \in \sethree $. 

\begin{assumption}[Prehensile {\it Pick}]
Each manipulator carries an end-effector that can immobilize the
object with a relative pose $ \grasp $ between the object and the
end-effector.
\label{ass:prehensile}
\end{assumption}

The manipulator is also able to stably \textit{place} the object.

\begin{assumption}[Object support]
The object can be in either: (a) \textit{Stable poses:} The object
lies in stable contact with a resting surface. This includes the
initial pose $ \pose_{init} $; or in (b) \textit{Picked poses:} Poses
of the object where one or more manipulators are carrying the object.
\label{ass:objectsupport}
\end{assumption}

The entire problem has a state space that is the Cartesian product of
all the constituent $\cspace$-spaces, i.e., $\taskspace
= \cspace \times \posespace_\object$. The collision-free subset is
defined as $ \tfree \subset \taskspace $.

\begin{assumption}[Singly-manipulable]
A single manipulator can pick up and move the object.
\label{ass:onearm}
\end{assumption}

Each object pose $\pose_{}$ where $o$ can be picked is associated with
at least one arm configuration $\state^i_{ {p}}$ for an $ \arm_i $,
which makes contact with $o$ with a pick $ \grasp $.

\begin{definition}[Hand-off]
\label{def:handoff}
A handoff is an instantaneous switching of an object from being
supported by a pick by $ \arm_i $ to being picked by $ \arm_j $, where
$ i\neq j $.
\end{definition}

A handoff pose constrains two manipulators. This means for a handoff
pose $ \pose_{hoff} $, there is an associated $
<\state^1_{pick}, \grasp^1> $ for $ \arm_1 $ and $
<\state^2_{pick}, \grasp^2> $ for $ \arm_2 $. Picks, places and
handoffs define transitions between \textit{modes} of the task.

\begin{assumption}[Problem Domain]
The following conditions hold: (i) 1 manipulator can reach the initial
pose of $ \object $, (ii) one manipulator can reach the final pose,
(iii) no stable poses of $ \object $ are reachable by both
manipulators.
\label{ass:domain}
\end{assumption}

\begin{definition}[Pick-and-place via Hand-off Problem]
Given an initial state
$(\state^1_{init}, \state^2_{init}, \pose_{init}) \in \tfree$ and a
final state
$(\state^1_{goal}, \state^2_{goal}, \pose_{goal}) \in \tfree$, a valid
solution to the \pdh\ problem is a path $ \Pi :[0,1]\rightarrow \tfree
$, such that $ \Pi(0) =
(\state^1_{init}, \state^2_{init}, \pose_{init}) $ and $ \Pi(1) =
(\state^1_{goal}, \state^2_{goal}, \pose_{goal})$, where $\exists\ 0<
t_{pick}<t_{hoff}<t_{place}\ < 1$, such that $ \Pi(t_{pick}) $ is a
valid picking state, then a handoff state $ \Pi(t_{hoff}) $ and a
valid placement $ \Pi(t_{place}) $.

\end{definition}

The cost function $ \cost: \Pi \rightarrow \mathbb{R}^+$ maps the path
to a real number. The experiments considered the duration of the
motion corresponding to $ \Pi $.

\begin{definition}[Asymptotically Optimal \pdh]
As computation time for an algorithm increases, the function
$ \cost(\Pi) $ of the solution $\Pi)$ discovered by the algorithm
asymptotically converges to the cost of the optimal solution $ \Pi^*$.
\end{definition}

\begin{figure}[t]
	\centering
	\includegraphics[width=0.48\textwidth]{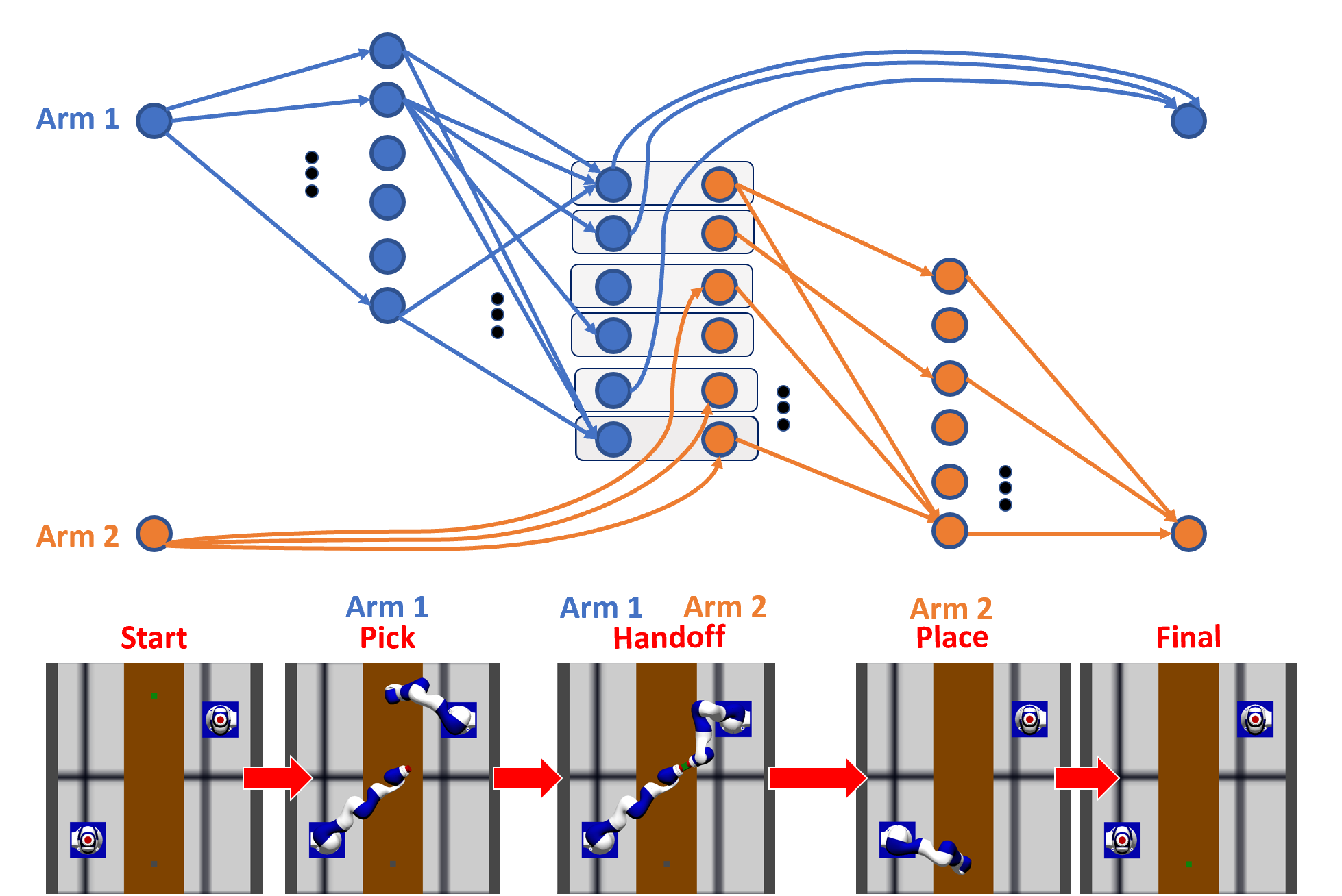}
	\vspace{-0.25in}

\caption{Mode graph $ \modegraph $ for the \pdh\ problem. The arms
start from the initial state on the left and progress through the
directed edges via picks, handoffs and places. Note that arm 2 moves
directly to handoffs from the initial state, and arm 1 directly moves
to the final configuration from the handoff. This introduces
underspecified intermediate states (picks and places). The frames
below show corresponding example states from a solution.}

	\vspace{-0.25in}
	\label{fig:mode_graph}
\end{figure}

\section{Method}
\label{sec:method}
This section first introduces critical tools used by the proposed
method and then goes over the  algorithmic steps. \vspace{-.05in}

\subsection{Components}

This following tools are used by the proposed framework.

\subsubsection{Transition Sampler}

A module generates picking (for the picking arm at the initial object
pose), handoff (for both arms and the object), and placement
configurations (for the placing arm at the final object pose). An
Inverse Kinematics solver is used for each arm at valid picking or
placing end-effector poses. A handoff state is generated by sampling a
new configuration for the arm that picks the object (for the
corresponding pick $g$), and then solving IKs for the handoff arm
given the object's pose. This sampling process will eventually include
all possible transitions.

\subsubsection{Mode Graph}

The mode graph $ \modegraph (\nodes_{\modegraph},
\edges_{\nodes_{\modegraph}})$ is a directed graph (Fig.
\ref{fig:mode_graph}), which contains all the components of the
\pdh\ problem and specifies the valid transitions between picks,
handoffs and places. Each node $ \node_{\modegraph} \in
\nodes_{\modegraph} $ contains \vspace{-.05in}
$$ <\state^1,\grasp^1>,<\state^2,\grasp^2> \ st. \ \state^1
\neq\ \mathtt{NULL}\ or\ \state^2
\neq\ \mathtt{NULL} \vspace{-.05in} $$ thereby representing
constraints for picks, places and handoffs. Picks are connected to
handoffs by a directed edge if the picking manipulator maintains the
same pick between its end-effector and the object. Similarly, edges
between handoffs and places exist if the placing transform matches the
pick at the handoff state for the same arm.

The construction of the graph ensures that a traversal through it for
both the arms satisfies the \pdh\ problem. Each edge can correspond to
multiple possible motion plans, which can reach different states in
the state space $\taskspace$. Multiple such states can satisfy an
underspecified mode constraint. For instance, while $ \arm_1 $ is
performing the pick, $ \arm_2 $ is free to proceed to some state that
might enable the next part of the solution to get optimized. The mode
graph $ \modegraph $ is input to planning method presented here.

\textit{Search over Mode Graph:} The initial $ (\state^1_{init},
\state^2_{init}, \pose_{init})$ and final states $ (\state^1_{goal},
\state^2_{goal}, \pose_{goal})$ correspond to two nodes on $
\modegraph $, denoted as $ \modeinit $ and $ \modegoal $. A traversal
of edges along the graph corresponds to executing motion plans for the
arms. A solution to the $ \pdh $ problem will then correspond to:
\vspace{-.05in}
$$ ( \ \modeinit, \node_\modegraph^i \cdots \node_\modegraph^j,  \modegoal ) \rightarrow \Pi \vspace{-.05in}$$

\subsubsection{Tensor Product Roadmap}
A roadmap $ \graph_1, \graph_2 $ is constructed for each constituent
manipulator. Then, a tensor product roadmap $ \mmgimp = \graph_1
\times \graph_2$ contains all combinations of vertices and
neighborhoods that exist in the constituent roadmaps
\cite{SoloveySH16:ijrr}. Prior work has shown that when the
constituent roadmaps are constructed with asymptotically optimal
properties, the tensor roadmap is also asymptotically optimal for the
multi-robot problem \cite{Dobson:2017aa, shome2019drrt}. The approach
does not need to explicitly store the tensor roadmap. Instead, the
search process implicitly explores $ \mmgimp $ online over the set of
constituent roadmaps $ \{\graph_1,\graph_2\} $, which is an input to
the algorithm. This allows searching a very dense structure without
incurring the space penalties.
\vspace{-.05in}

\subsection{Algorithm}

The proposed $ \mmdrrtstar $ simultaneously searches over \textit{(a)}
different configurations $ \State \in \cspace$ of the arms as
expressed via $ \mmgimp $, and \textit{(b)} different poses of the
object. A node in the search tree
$ \tree $ keeps track of the composite arm configuration $
\State $, the mode $ \mode $ and fully specifies an object pose as it is
the result of manipulation operations such as picks, handoffs and
places. The method does so by building a search tree over the tensor
roadmap and the mode graph $ \modegraph $.  For notational simplicity,
the object pose is not explicitly outlined in the algorithmic
description, but is fully specified given a specific tree node given
the sequence of arm configurations and modes along that branch of the
tree.

The key features of the proposed method is the principle of searching over this decomposed search space of individual arm roadmaps, and the mode graph concurrently to effectively search for solutions in the task space $ \taskspace $. Following the benefits shown in previous work (\drrtstar), the search over this representation proves efficient when coupled with heuristic guidance of node expansions of the search tree. At each step, the algorithm tries to make progress from a node which describes an arm configuration, and an object pose associated with a mode $ \mode $. The $ \modegraph $ expresses goals for the adjacent modes and can be used to guide the expansion towards them for each arm. In the case of underspecified modes, some of the arms might not have a specific target. In this case the guidance is used to look ahead to modes on $ \modegraph $ that might be further away (e.g., at the picking mode, there is a goal for the
manipulator that will pick the object but the handoff manipulator
needs to make progress towards the handoff ``region''). This strategy scales to multiple manipulators and modes. Such a search strategy lets the proposed method quickly make progress across modes, while ensuring both different arm motions, as well as different transitions are explored over time.

\begin{figure}[t]
\centering
\includegraphics[width=0.48\textwidth]{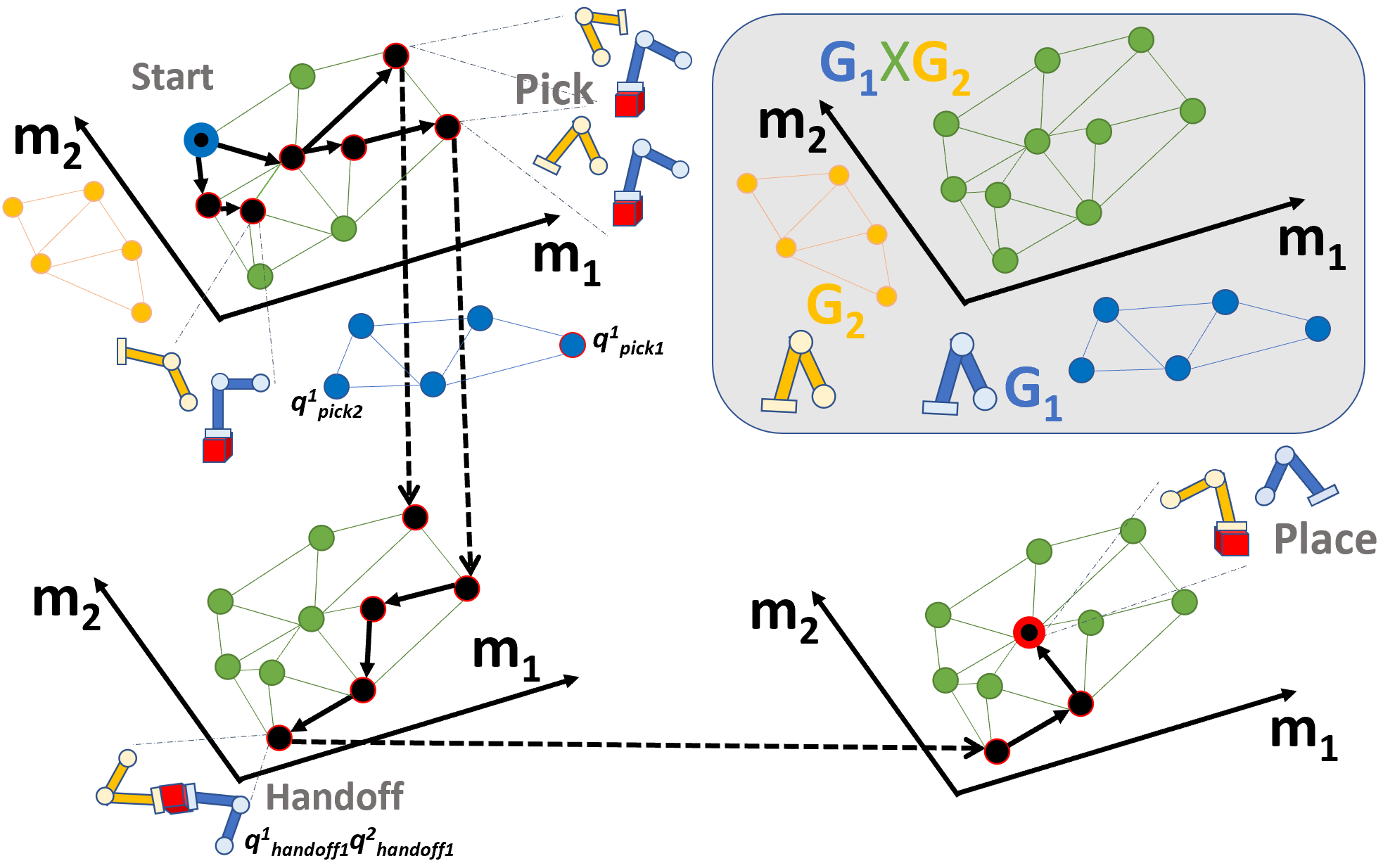}
\vspace{-0.3in}
\caption{\textit{(Top right):} The tensor product roadmap $
  \graph_1\times\graph_2 $ (green) contains all combinations of nodes
  and edges from the constituent roadmaps for the arms $m_1$ (blue)
  and $m_2$ (yellow). \textit{(Rest)}: The tree expansion for solving
  the \pdh\ problem starts at the top left. Three different picking
  states are shown, where two of them arise from the same pick
  $\state^1_{pick1}$ for $m_1$. The tree explores different options
  for the unconstrained arm $m_2$ in these cases. The dashed arrows
  indicate mode changes, which lead to handoff and eventually place
  states, while implicitly searching for arm motions over the tensor
  roadmap.}
\vspace{-0.25in}
\label{fig:tree}
\end{figure}

The high-level method is described in Algorithm
\ref{algo:drrtstar}. The mode graph, the initial and target modes are
inputs.  The
algorithm builds a tree $ \tree $ where each node of the tree is
composed of the configuration of the manipulators $ \State $ and the
mode $ \mode $, which is sufficient to represent a full task space state. Lines 1-4 initialize the algorithm with the initial
starting configuration in $ \modeinit $. The loop over Lines 5-9
expands the tree, updates the path $ \Pi $ if found, and keeps track
of the best solution discovered so far.

\vspace{-0.15in}
\begin{algorithm}[!ht]
\caption{$\mmdrrtstar {\tt(} \modegraph, \modeinit, \modegoal, \tensorset {\tt)}$}
\label{algo:drrtstar}
$\Pi_{\textup{best}} \gets \emptyset$\;
$ \State^{init} \leftarrow (\modeinit.\state^1, \modeinit.\state^2)$\;
$\tree.{\tt init}(<\State^{init}, \modeinit>)$\;
$\vlast \gets <\State^{init}, \modeinit>$\;
\While{${\tt time.elapsed}() < {\tt time\_limit}$}
{       
    $\vlast \gets {\tt Expand\_\mmdrrtstar(} \tree, \vlast, \modegraph, \tensorset {\tt)}$\;
    $\Pi \gets {\tt Connect\_to\_Target(} \tensorset, \tree, \modegoal {\tt)}$\;
    \If{$\Pi \neq \emptyset\ \cap\ \cost(\Pi) < \cost(\Pi_{\textup{best}})$}
    {
        $\Pi_{\textup{best}} \gets {\tt Trace\_Path(} \tree, \modegoal {\tt)}$
    }
    
}
{\bf return $\Pi_{\textnormal{best}}$}
\end{algorithm}
\vspace{-0.2in}

The expansion of the tree $ \tree $ per iteration is described in
Algo. \ref{algo:drrtstar_expand}. $\vlast $ keeps track of whether the
greedy heuristic allows the tree to approach the goal and continues
the greedy behavior if it does so. Lines 1-3 describe the random
exploration behavior which randomly selects a node in the tree,
specifying the configuration $ \Snear $ and mode $ \mnear $. A random
neighbor $ \Snew $ for the configuration $ \Snear $ is selected from $
\mmgimp $. Note that this can be performed by randomly selecting an
adjacent vertex in each $ \graph_i $ that composes $ \mmgimp $.

\begin{figure*}[t!]
	\centering
	\includegraphics[width=0.4\textwidth]{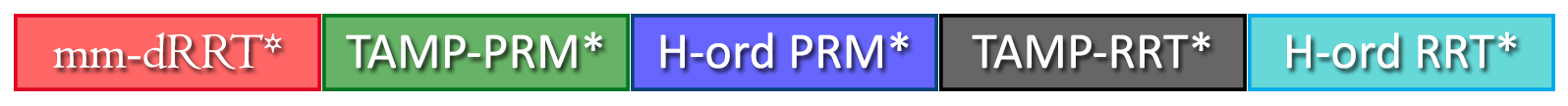}\\
	\includegraphics[width=0.25\textwidth]{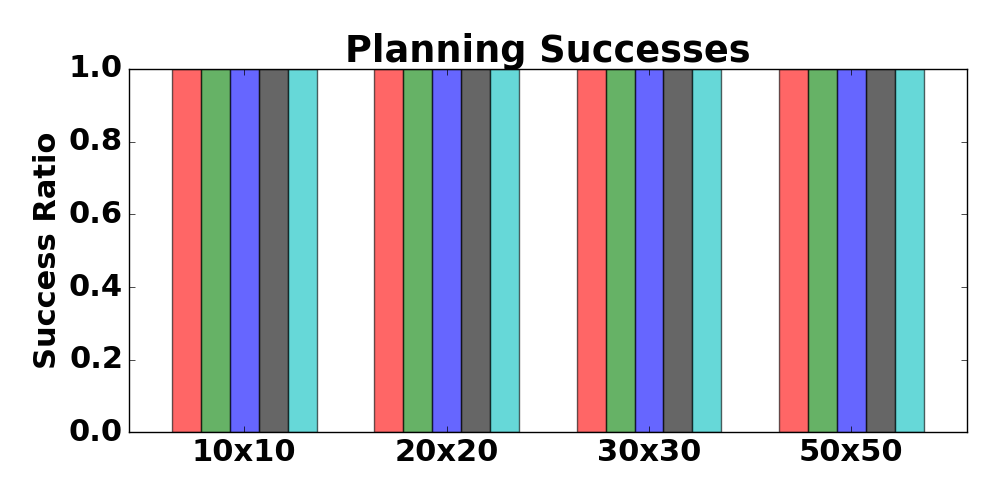}
	\includegraphics[width=0.25\textwidth]{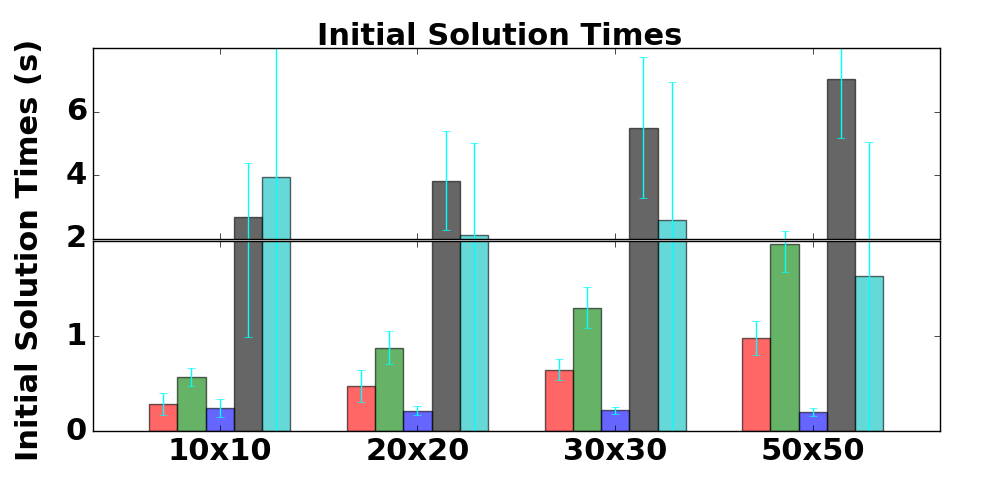}
	\includegraphics[width=0.25\textwidth]{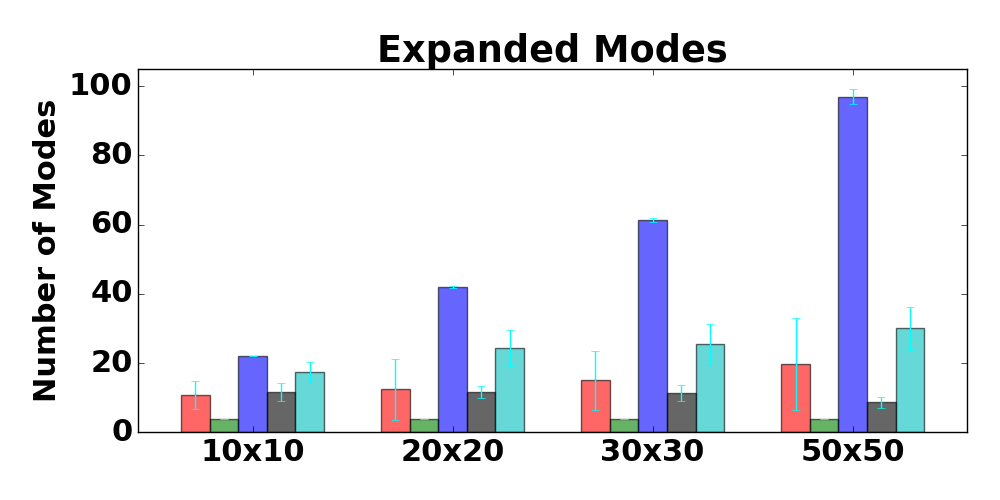}\\
	\includegraphics[width=0.14\textwidth]{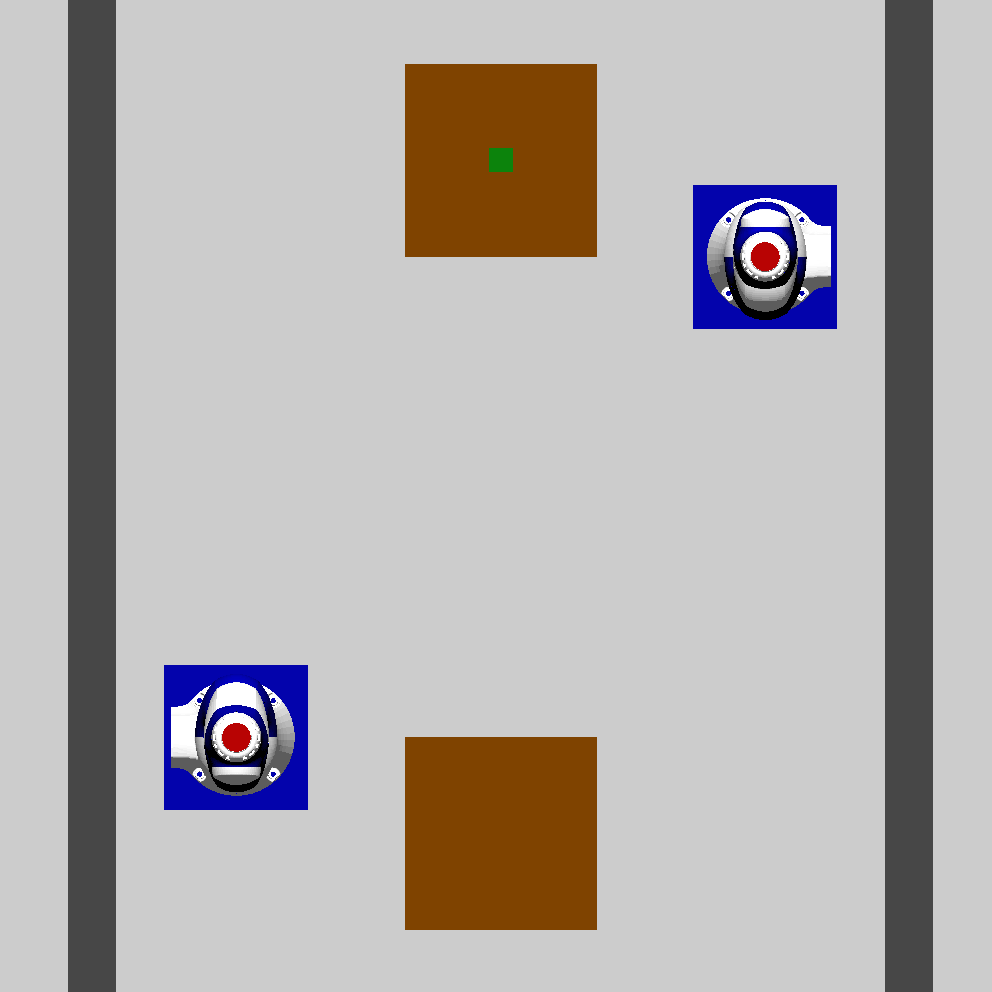}
	\includegraphics[width=0.64\textwidth]{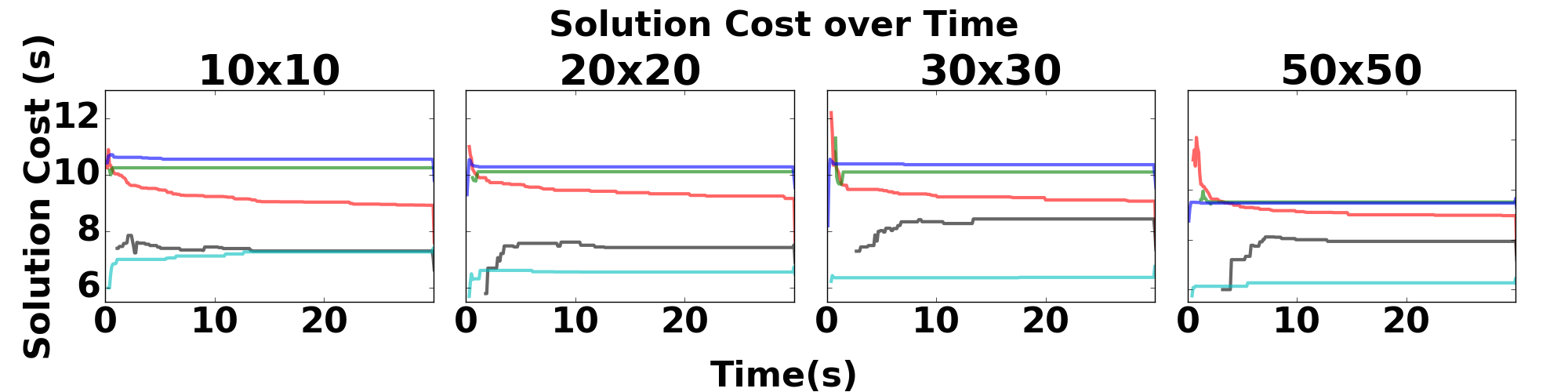}
	\vspace{-0.1in}
	\caption{Results for the tabletop (shown at
          \textit{bottom left}) for different sizes of $ \modegraph $;
          \textit{top left}: planning success ratio, \textit{top
            middle}: average initial solution times; \textit{top
            right}: average number of expanded modes after 30s;
          \textit{bottom right}: The average solution cost over time
          for the methods. }
	\vspace{-0.25in}
	\label{fig:tabletop_benchmark}
\end{figure*}

The greedy behavior in Lines 4-6 invokes an oracle $ \ioracle $, which
biases the tree expansion towards an adjacent mode. If a new node is
added to the tree in Lines 7-8, the subroutine $ {\tt
  Add\_and\_Rewire\_Neighborhood} $ performs collision checking to
verify whether the node addition is valid. The rewiring refines the
neighborhood of the added node, taking care of respecting mode
constraints, i.e., only rewiring nodes belonging to the same
transition between modes. Lines 9-12 checks whether the newly added
node satisfies the constraints in an adjacent mode $ \mode \in
\mathcal{N}_\modegraph $. If so, the adjacent mode is added to the
tree and heuristic gradient following is triggered. Lines 14-15 also
triggers the greedy behavior if the added node displays a better
heuristic $ \heuristic $.

\vspace{-0.15in}
\begin{algorithm}[ht]
\caption{${\tt Expand\_\mmdrrtstar(}  \tree, \vlast, \modegraph, \tensorset {\tt)}$}
\label{algo:drrtstar_expand}
\If{$\vlast = \emptyset$}
{
    $\vnear = <\Snear, \mnear> \gets {\tt select(} \tree {\tt)}$\;
    $\vnew = <\Snew, \mnear> \gets {\tt Random\_Neighbor (} \Snear, \tensorset {\tt)}$\;
}
\Else
{
    $\vnear = <\Snear, \mnear> \gets \vlast$\;
    $\vnew = <\Snew, \mnear> \gets {\tt \ioracle (} \vnear, \tensorset, \modegraph {\tt)}$\;
}

\If{ $\vnew \notin \tree$ }
{
$\tree.{\tt Add\_and\_Rewire\_Neighborhood(} \vnew {\tt)}$\;
}

$\modenbor \gets {\tt Adjacent\_Modes}( \modegraph, \mnew)$\;
\If{$ {\tt Satisfies(} \Snew, \mode{\tt )} $ \bf{for} $ \ \mode \in \modenbor$}
{
	$\vnew \gets <\Snew, \mode>$\;
	$\tree.{\tt Add\_and\_Rewire\_Neighborhood(} \vnew {\tt )}$\;
	${\bf return}\ \vnew$\;
}
\If{${\tt \heuristic}(\vnew) < {\tt \heuristic}( {\tt Parent(}  \vnew {\tt )} )$ }
{
  ${\bf return}\ \vnew$\;
}
\lElse
{
    ${\bf return}\ \emptyset$
}
\end{algorithm}
\vspace{-0.2in}

The greedy node generation is outlined in Algo.
\ref{algo:idrrtstar_oracle}. This module is aware of the mode graph $
\modegraph $. Line 1 decomposes the configuration $ \Snear $ into the
constituent manipulator configurations $ (\snear_i) $. The subroutine
$ {\tt Traverse\_and\_Get\_Targets(} \mnear, \modegraph {\tt )} $
performs a search over the mode graph to find a set of grounded
configuration targets for each manipulator. Lines 3-4 chooses a
neighbor in $ \mmgimp $ that minimizes the heuristic $ \heuristic $ to
target configurations $ T_i $. The new node $ \vnew $ is recomposed in
Lines 5-6.

\vspace{-.15in}
\begin{algorithm}[!h]
\caption{${\tt \ioracle (} \vnear, \tensorset, \modegraph {\tt)}$}
\label{algo:idrrtstar_oracle}
$ ( \snear_i ) \gets \Snear$\;
$ ({T}_i) \gets {\tt Traverse\_and\_Get\_Targets(} \mnear, \modegraph {\tt )} $\;
\For{$ i : 1 \rightarrow {\tt Num\_Robots(} \tensorset {\tt )} $}
{
	{

		$ \snew_i \leftarrow \underset{x\in{\tt Adj}(\snear_i,\graph_i)}{argmin}{\tt \heuristic(} x, {T}_i, \graph_i {\tt )} $ \; 
	}
}

$\Snew \gets (\snew_i)$\;
$\vnew \gets < \Snew, \mnear >$\;

\textbf{return} $ \vnew $

\end{algorithm}
\vspace{-.2in}

\noindent\textit{Implementation Details}: The heuristic function $ \heuristic $ used is the \textit{MAKESPAN} or
maximum of distances over all the manipulators, scaled by the maximum
velocities of the joints. This lets the measure be consistent with $
\cost $ that describes the duration of $ \Pi $. Tree node additions
are branched-and-bounded. The subroutine $ {\tt
  Traverse\_and\_Get\_Targets(} \mnear, \modegraph {\tt )} $ is aware
of the heuristic estimates over $ \modegraph $ and selects from the
adjacent modes appropriately. The heuristic information over each $
\graph_i $ in the form of all-pairs shortest paths in each roadmap is
precomputed and cached. Each time a new mode is expanded, the
corresponding configuration can be added the constituent roadmap $
\graph_i $. The heuristic information can be approximated by the
nearest valid roadmap neighbor. The heuristic between two
configurations not yet part of the roadmap (i.e., parts of the $
\modegraph $ not yet explored) are estimated by pairwise
\textit{MAKESPAN}. Goal biasing involves a fraction of iterations
where $ {\tt Select} $ chooses from tree nodes, which have made the
most progress over $ \modegraph $.

\section{Sketch of Properties}

Given a mode graph $ \modegraph $, an optimal solution to the
\pdh\ problem $ \Pi^* $ will trace a sequence of task space states $
(\State^*_{pick},\pose_{init}) $, $
(\State^*_{handoff},\pose^*_{handoff}) $, and $
(\State^*_{place},\pose_{goal}) $ that represent the mode transitions
that solve the problem optimally. In order to ensure asymptotic
optimality, given enough samples and time, the algorithm has to
guarantee the following asymptotically: \textit{(a)} it discovers a
sequence of transitions that converges to the sequence of optimal mode
transitions, and \textit{(b)} it discovers a collision free path
between each pair of mode transitions, that converges to the optimal
such connection.

Firstly using results from \drrtstar\cite{Dobson:2017aa,shome2019drrt}
it can be argued that given any pair of states that might represent a
pair of mode transitions, the tree expansion strategy in Algo.
\ref{algo:drrtstar_expand} ensures asymptotic optimality of the
connection between them given enough time and large enough number of
samples in the underlying roadmaps.

Secondly, the problem of discovering the optimal sequence of
transitions has to be addressed. In cases where the mode transition
states are fully specified in $ \modegraph $, like for handoff states,
their discovery is ensured if the algorithm guarantees exploration of
all modes in $ \modegraph $.  Line 2,3 of Algo.
\ref{algo:drrtstar_expand} guarantees this exploration. Nevertheless,
some of the modes in $ \modegraph $, say picks, underspecify the
constraints on the full state. The exploration of different options
for an unconstrained arms have to guarantee convergence to such a
state, say $ \State^*_{pick} $ in the case of picks. Line 2,3 of Algo.
\ref{algo:drrtstar_expand} again guarantees that every node in the
tensor roadmap is explored. As the size of the constituent roadmaps
increases, asymptotically the existence of a sample close enough to $
\State^*_{pick} $ is guaranteed to exist in the tensor roadmap,
ensuring convergence.

Bringing both the arguments together, $ \mmdrrtstar $ is AO given
enough iterations and enough samples in the individual roadmaps for
solving the $ \pdh $ problem over an input $ \modegraph $. The
indications also imply that if $ \modegraph $ keeps getting augmented
the arguments can extend to general planning problems that can be
described by such a mode graph. The detailed exposition of the
corresponding proof is left for future work.

\section{Results}
\label{sec:results}
This section evaluates the proposed method in simulated setups
involving multiple robotic arms.  The focus in three considered
benchmarks is on the following aspects:

\begin{myitem}
\item[-] \textit{Size of $ \modegraph $}: The experiments consider an
  input mode graph $ \modegraph $ of increasing size and connectivity.
\item[-] \textit{Tight Spaces}: Obstacles are introduced in the second
  benchmark, which make the problem harder.
\item[-] \textit{Scalability}: The effect of planning for an
  increasing number of high \dof\ robots.
\end{myitem}

The benchmarks use $ 7$-$\dof $ \textit{KUKA-iiwa14} arms placed in an
offset manner so that they can execute handoffs, while one of them can
pick the object at the initial pose and another robot can place it at
the goal pose (e.g., as shown in Fig \ref{fig:tabletop_benchmark}).

\begin{figure*}[ht]
	\centering
	\includegraphics[width=0.4\textwidth]{figures/images/labels}\\
	\includegraphics[width=0.25\textwidth]{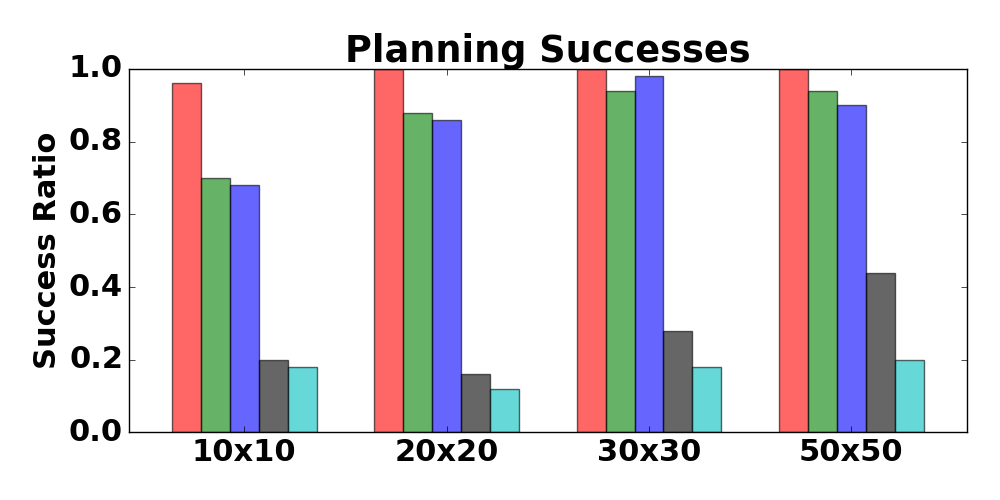}
	\includegraphics[width=0.25\textwidth]{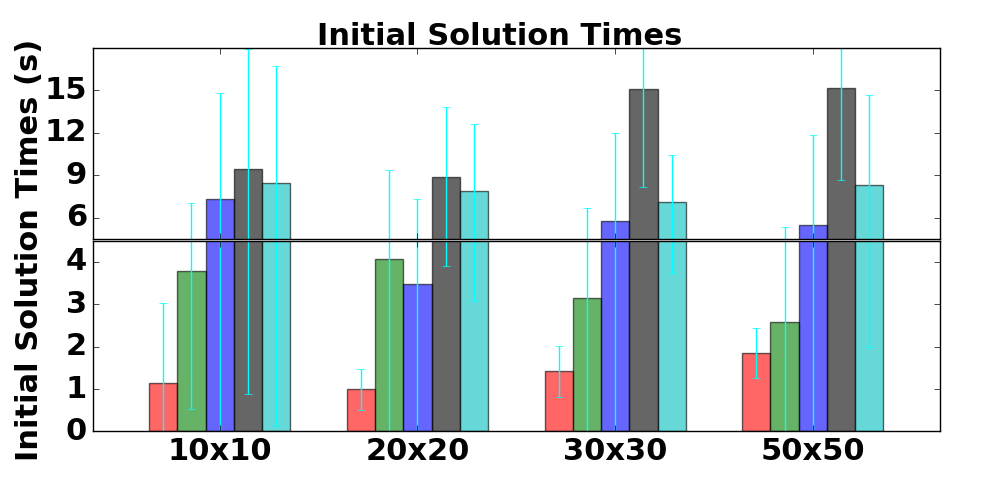}
	\includegraphics[width=0.25\textwidth]{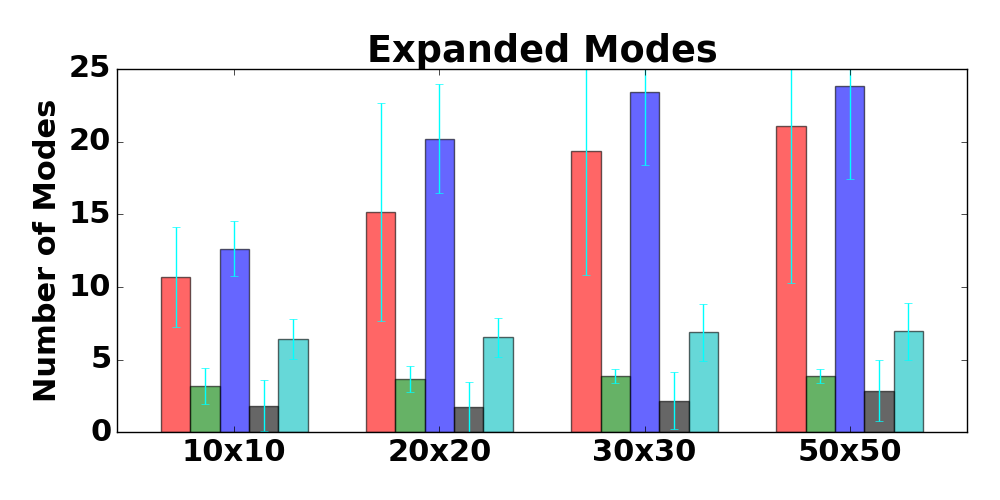}\\
	\includegraphics[width=0.14\textwidth]{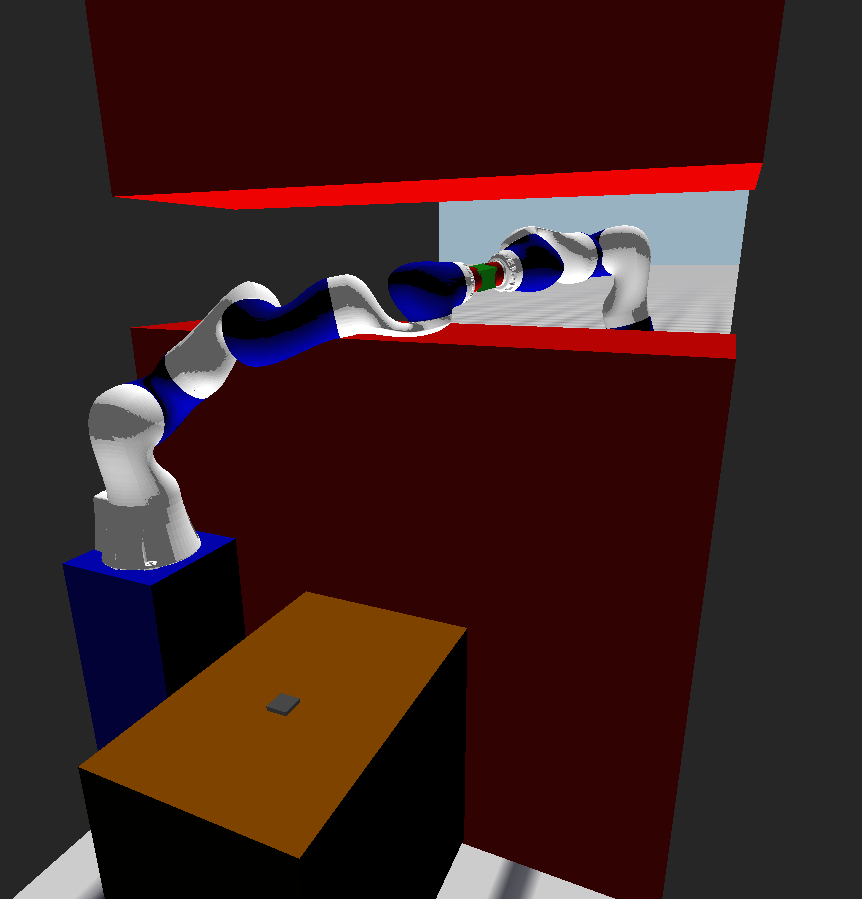}
	\includegraphics[width=0.64\textwidth]{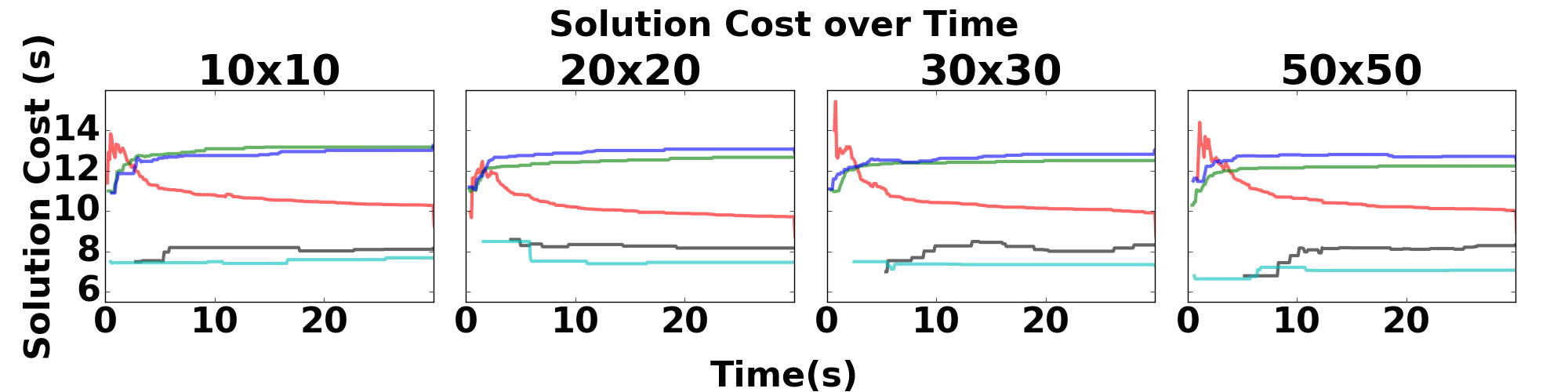}
	\vspace{-0.1in}
	\caption{Narrow passage (\textit{bottom left}) results
          reported for different sizes of $ \modegraph $; \textit{top
            left}: planning success ratio, \textit{top middle}:
          average initial solution times; \textit{top right}: average
          number of expanded modes after 30s; \textit{bottom right}:
          The average solution cost over time for all the methods.}
	\vspace{-0.15in}
	\label{fig:harder_benchmark}
\end{figure*}

\begin{figure*}[t!]
	\centering
	\includegraphics[width=0.13\textwidth]{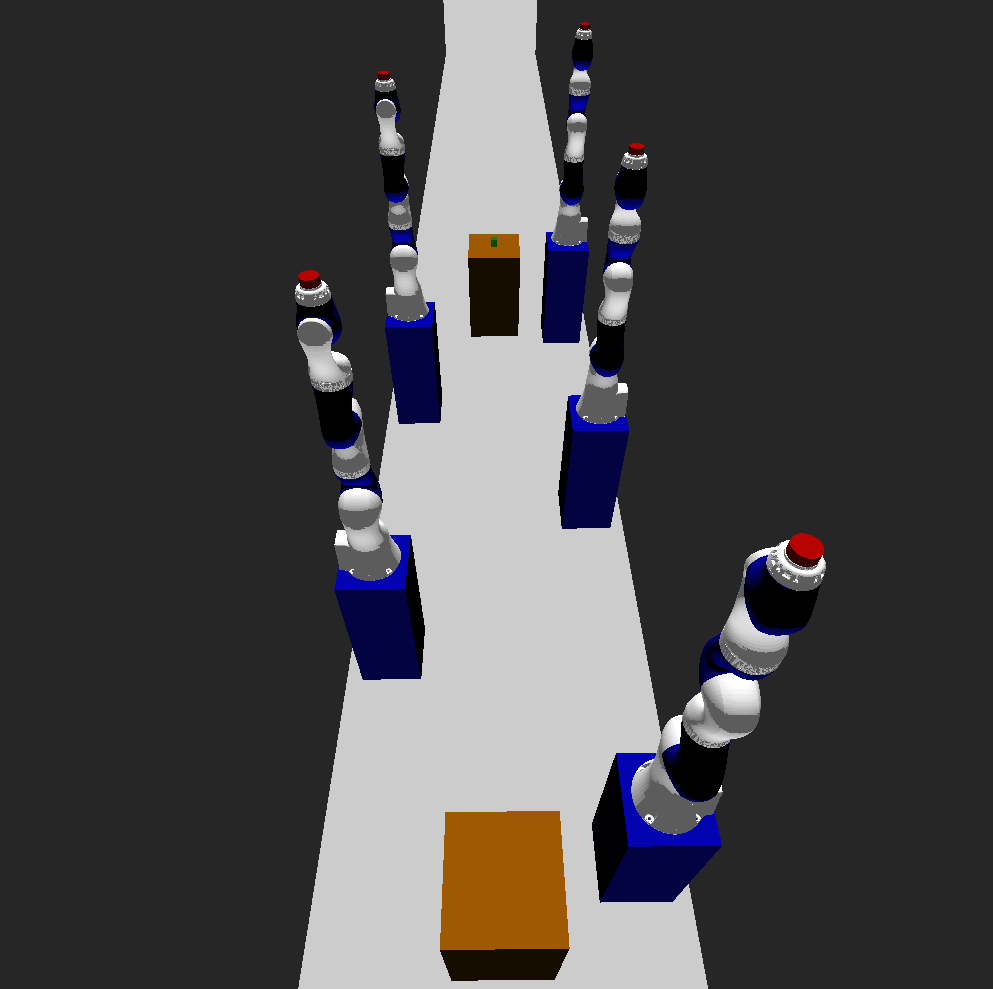}
	\includegraphics[width=0.23\textwidth]{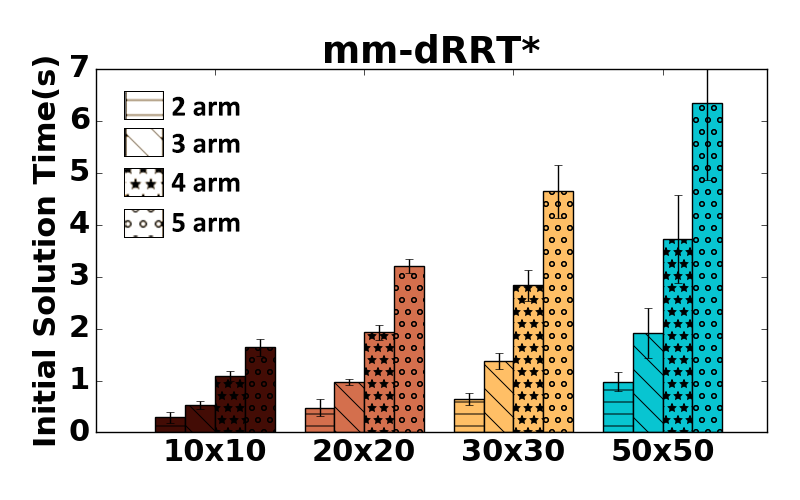}
	\includegraphics[width=0.23\textwidth]{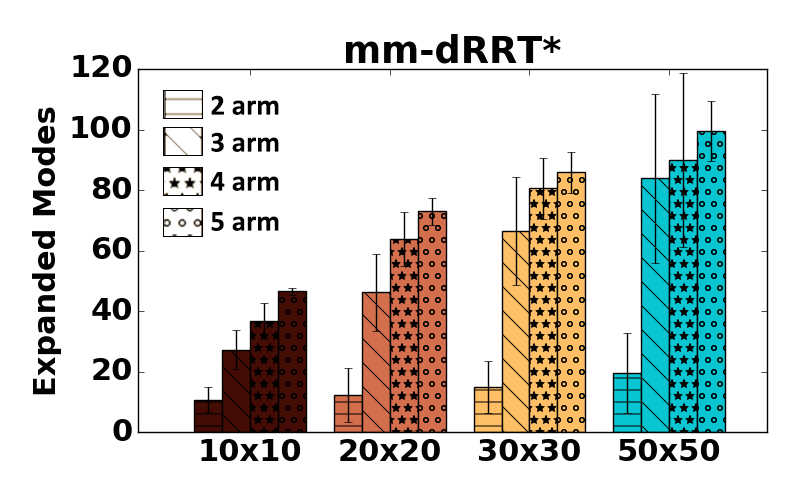}
	\includegraphics[width=0.36\textwidth]{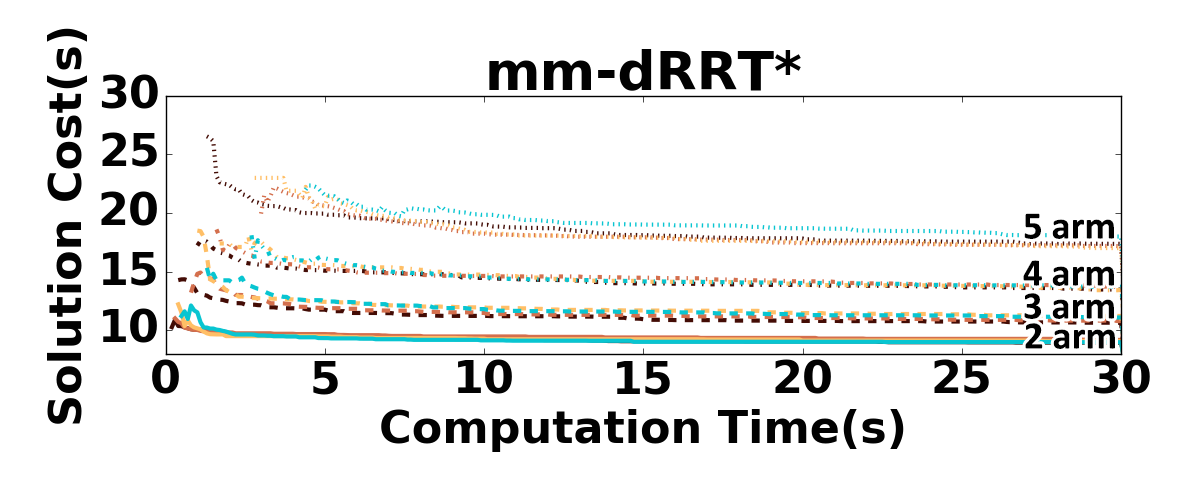}
		\vspace{-0.1in}
	\caption{N-arm chained handoff (\textit{left}) demonstration results; \textit{second:} average initial solution times for different sizes of $ \modegraph $; \textit{middle}: average number of expanded nodes after 30s; \textit{right}: Average solution cost over time for different sizes of $ \modegraph $.  }
		\vspace{-0.25in}
	\label{fig:three_arm}
\end{figure*}

The problem of generating candidate picks is not the focus in this
work. The benchmarks deal with a cuboid object for which a single
picking pose is defined at the top face and one at the bottom
face. Each handoff configuration involves the manipulators picking the
object on opposite faces. The increase in the size of $ \modegraph $
is achieved by solving for an increased number of inverse kinematics
solutions for the pick and handoff configurations. The target object
orientation is flipped to allow the placing manipulator to properly
place the object. The number of the arm picking and handoff
configurations range over $s = 10,20,30 $ and $ 50 $ for each
benchmark, leading to $ s^n $ combination of choices for pick and
handoff orderings for $ n $ arms. These configurations are collision
checked to ensure that they are valid for each scene.

The proposed $ \mmdrrtstar $ is compared against two different
strategies of \tamp\ frameworks, each one of them using two different
underlying AO motion planners.

\noindent (a) \textit{TAMP-\prmstar, TAMP-\rrtstar}: The \textit{TAMP}
variants search sequentially over the space of modes. They start with
$\modeinit $ and first plan a solution by considering all the adjacent
$\mode$ to $\modeinit $ (i.e., the picking states). Underspecified
manipulators are trivially assigned to a home position. The method
then commits to the reached $ \mode $ and the search progresses till $
\modegoal $ is reached. TAMP-\rrtstar\ is rerun from the start if
there is any time remaining.

\noindent (b) \textit{H-ord \prmstar, H-ord \rrtstar}: The
heuristically ordered variants perform a DFS over $\modegraph $. The
approach similarly considers a set of target configurations along
adjacent modes. These are then heuristically ordered and tried in
sequence. Once one succeeds, the search progresses until $ \modegoal $
is reached or planning fails. The method then backtracks and keeps
trying till the time limit.

\noindent (b) \textit{H-ord \prmstar, H-ord \rrtstar}: The
heuristically ordered variants first select the best adjacent neighbor
for every mode on $ \modegraph $, in terms of the heuristic estimate,
starting from $ \modeinit $. They solve motion planning queries till
they find the solution. If they fail at any point, they move on to the
next best neighbor according to the heuristic. So as to ensure a fair
chance to all modes, a time budget of 10s is assigned on each
individual query. The heuristic is the pairwise \textit{MAKESPAN}
over $ \modegraph $ scaled by the max velocity of the arms.

Each experiment was executed single-threaded on a 32 core Intel(R)
Xeon(R) CPU E5-1660 v3 @ 3.00GHz machine with 32 GB of RAM. Data is
reported as averages over 50 randomized trials of 30s for different $
\modegraph $. Individual arm roadmaps of size 200 were used for $
\mmdrrtstar $ (giving rise to an implicit tensor product of $200^n$
where $n$ is the number of arms. The \prmstar\ variants were executed
on a 20k node roadmap constructed over the 14-dim configuration space
of both robots. The implicit nature of the tensor roadmap allows for
two-orders of magnitude benefits in space requirements relative to
\prmstar.

\noindent\textbf{Tabletop Benchmark}:
Fig. \ref{fig:tabletop_benchmark} shows the runs of the algorithm on a
tabletop scene. The data indicates the baseline performance of the
algorithms in searching over different sizes of $ \modegraph $ when
the planning problem is easy. $ \mmdrrtstar $ finds better quality
solutions than \prmstar\ variants. \prmstar\ is the fastest because it
has to do very little online computation, but $ \mmdrrtstar $ is
competitive in terms of initial solution times. \rrtstar\ takes longer
to find better solutions.
 
\noindent\textbf{Narrow Passage Benchmark}: A dividing wall with a
slit is introduced to the scene as shown in Fig.
\ref{fig:harder_benchmark}. \prmstar\ now suffers in success rate from
the brittleness of the roadmap, and takes longer to find the initial
solution. \rrtstar\ also suffers in terms of success
rate. \mmdrrtstar\ succeeds in all but one run and its initial
solution time is better than competing methods over all runs. The
number of expanded modes indicates improved exploration over
$\modegraph$ for \mmdrrtstar.

\noindent\textbf{N-arm Scalability}: Fig \ref{fig:three_arm} shows the
results from a execution of the algorithm for a chained sequence of
handoffs involving 2, 3, 4 and 5 arms, to transfer the object across
the workspace. Fast initial solution times indicate that the method
scales well with a larger number of robots. For 5 robots and 50 picks
and handoffs, the planner simultaneously searches over a tensor
roadmap of size $ 200^5 $, and $ 50^5 $ possible mode traversals.

\section{Discussion}
\label{sec:discussion}
The current work proposes a multi-modal \drrtstar approach with a
specific focus on solving a pick-and-place via handoff problem for
high-\dof manipulators. Results indicate that the principles of tensor
roadmap decomposition and heuristic guidance transfer nicely to the
multi-modal domain. The proposed anytime method finds fast initial
solutions and is robust to harder instances of the
problem. Theoretical arguments indicate asymptotic optimality of the
approach for a given set of mode transitions.

The promising results motivate applying the same framework to more
general problem classes of multi-arm \tamp\ involving richer sets of
modes and operations, which can include like generalized grasps,
online grasp discovery, regrasping, bi-manual manipulation,
multi-object rearrangement, and tasks involving dynamics, such as
throwing, pushing and within-hand manipulation.

{
\bibliographystyle{IEEEtran}

}

\end{document}